\documentclass[10pt,twocolumn,letterpaper]{article}
\usepackage{amsmath}
\usepackage{amsfonts}
\usepackage{amssymb}
\usepackage{algorithm}
\usepackage{algorithmic}
\usepackage{multirow}
\usepackage{pifont}
\usepackage{colortbl}
\usepackage[pagenumbers]{iccv} 

\definecolor{iccvblue}{rgb}{0.21,0.49,0.74}
\usepackage[pagebackref,breaklinks,colorlinks,allcolors=iccvblue]{hyperref}

\title{Marine Saliency Segmenter: Object-Focused Conditional Diffusion with Region-Level Semantic Knowledge Distillation}

\author{
    Laibin Chang$^{1}$, Yunke Wang$^{2}$, JiaXing Huang$^{3}$, Longxiang Deng$^{1}$, Bo Du$^{1}$, Chang Xu$^{2}$ \\
    $^{1}$ School of Computer Science, Wuhan University \\
    $^{2}$ School of Computer Science, The University of Sydney \\
    $^{3}$ School of Computer Science and Engineering, Nanyang Technological University\\
}

\begin{document}
\maketitle
\begin{abstract}
    Marine Saliency Segmentation (MSS) plays a pivotal role in various vision-based marine exploration tasks. However, existing techniques often face the dilemma of imprecise boundaries due to the interference-rich nature of underwater environments, where suspended particles, low contrast, and color distortion hinder accurate segmentation. Meanwhile, despite the impressive performance of diffusion models in visual tasks, there remains potential to further leverage contextual semantics to enhance feature learning of region-level salient objects, thereby improving segmentation outcomes. Building on this insight, we propose DiffMSS, a novel marine saliency segmenter based on the diffusion model, which utilizes semantic knowledge distillation to guide the detection of marine salient objects. Specifically, we design a region-word similarity matching mechanism to identify salient terms at the word level from the text descriptions. These high-level semantic features guide the conditional feature learning network in generating salient and accurate diffusion conditions with semantic knowledge distillation. To further refine the segmentation of fine-grained structures in unique marine organisms, we develop a dedicated consensus deterministic sampling to suppress overconfident missegmentations. Extensive experiments demonstrate the superior performance of DiffMSS over state-of-the-art methods in both quantitative and qualitative evaluations.
\end{abstract}

\section{Introduction}\label{Introduction}
\begin{figure}[!tp]
    \setlength{\abovecaptionskip}{0.1cm}
    \setlength{\belowcaptionskip}{-0.2cm}
    \centering
    \includegraphics[width=0.456\textwidth]{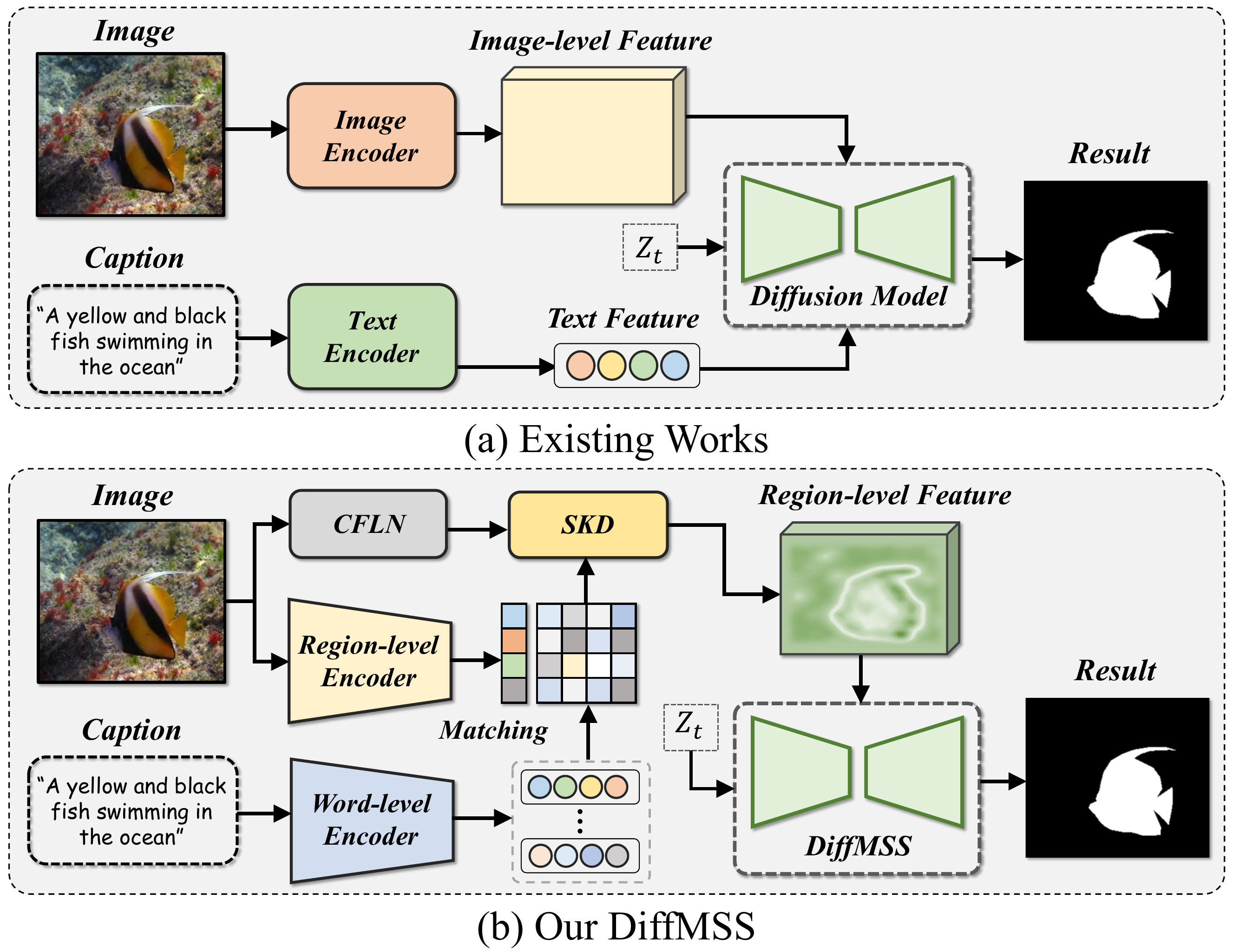}
    \caption{Different from existing diffusion-based methods that directly condition on coarse-grained image-level visual or text features, our DiffMSS designs Region-Word Matching with Conditional Feature Learning Network (CFLN) and Semantic Knowledge Distillation (SKD) to capture fine-grained region-level visual features as accurate conditions for object-focused diffusion.}
    \label{1-0The_First_Figure}
    \vspace{-0.2cm}
\end{figure}
Marine Saliency Segmentation (MSS) focuses on segmenting visually salient objects within complex underwater environments to meet the growing requirement for fine-grained object recognition \cite{Zheng2024Coralscop}. Functionally, accurate recognition of marine instances contributes to applications like organism identification \cite{Li2024Organism}, autonomous navigation \cite{Manzanilla2019Autonomous}, and object detection \cite{Chen2024Cwscnet}. However, the raw images captured directly by underwater vehicles tend to lose visual saliency, presenting various types of degradation, such as color distortion, low contrast, and blurred details \cite{Zheng2025Marineinst}. Underwater degraded images with these defects usually exhibit indistinguishable object boundaries and a camouflaged appearance.

With advances in large-scale annotated datasets \cite{Fan2022salient, zhang2025Referring} and deep network architectures, many saliency detection methods \cite{Yin2024Camoformer, Li2022Saliency_Dec, Tang2024Chain} in natural image domains have made remarkable performances. However, they face challenges in underwater environments, where the poor visibility and fine-grained structures of marine organisms (\textit{e.g.}, fish, corals) greatly degrade the accuracy \cite{Zheng2025Hkcoral}.
Existing MSS methods \cite{Deng2023RMFormer, Islam2020SUIM, Islam2021SVAM, Fu2024MASNet, Lian2023Watermask} follow the basic paradigm of a learning-based backbone and a decoder for segmentation. However, they usually stack multiple convolutional sequences with limited representational power into deep networks to extract deep features that require a lot of computational resources \cite{Hong2023USOD10k, Du2023USD}. Without well-designed backbones, they remain vulnerable to visual degradation and suffer from inaccurate boundary segmentation.

Given the specific challenges posed by the MSS task, we explore the diffusion model \cite{Song2020DDIM} as a fitting solution due to its strong generative capabilities. Despite the impressive performance of diffusion models \cite{Pnvr2023LD_ZNet, Chen2024Hidiff} with conditional prompts in common segmentation tasks, the potential of leveraging contextual semantics to generate the diffusion conditions remains underexplored. Moreover, these models usually adopt coarse-grained image-level or text-level features as conditions for diffusion (as shown in Fig. \ref{1-0The_First_Figure}). While image-level captions provide contextual information, word-level concepts can offer more precise semantic cues related to salient regions \cite{Wu2024Co_Decomposition}. By extracting and aligning these word-level semantics with visual features, we can guide the diffusion model to focus on these key salient regions and improve its segmentation accuracy.

\begin{figure*}[!htp]
	\setlength{\abovecaptionskip}{0.1cm}
	\setlength{\belowcaptionskip}{-0.1cm}
	\centering
	\includegraphics[width=1\textwidth]{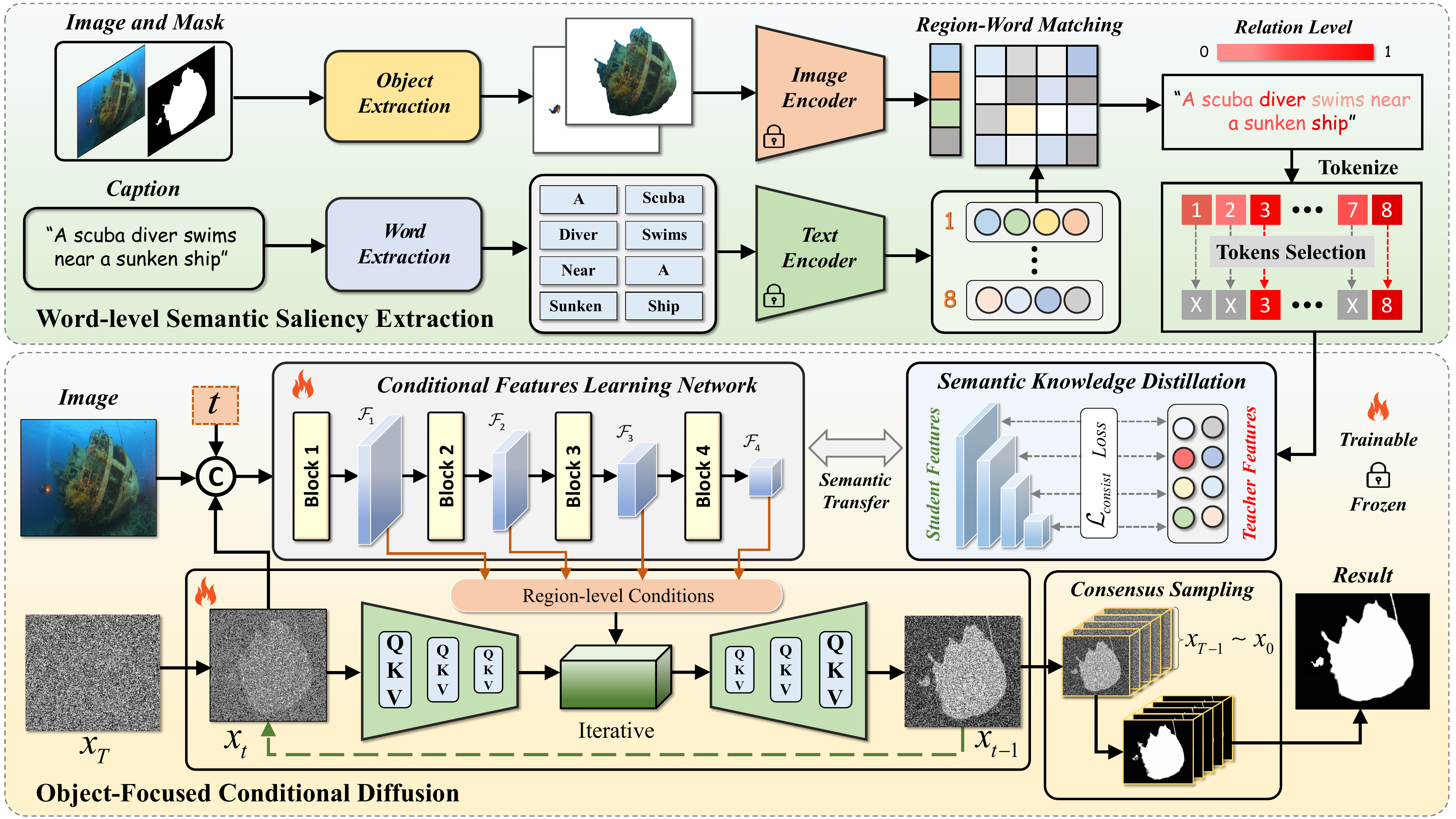}
	\caption{
    Overview of the proposed DiffMSS. Its training mainly contains three procedures: 
    (a) Given the image caption, Word-level Semantic Saliency Extraction identifies salient concepts in terms of words via region-word matching.
    (b) Semantic Knowledge Distillation transfers the identified word-level semantic tokens into the Conditional Feature Learning Network to generate region-level conditions for object-focused diffusion.
    (c) Consensus Sampling enables fine-grained structural segmentation of intricate marine instances via deterministic ensemble scheme. 
    Note the top green box (i.e., Word-level Semantic Saliency Extraction) and the Semantic Knowledge Distillation in the bottom yellow box are utilized exclusively during training and will deactivated in the testing phase.
    }
	\label{1-1The_Overall_Framework}
    \vspace{-0.3cm}
\end{figure*}

Motivated by the aforementioned analysis, we propose DiffMSS, an innovative diffusion-based marine saliency segmenter that leverages a region-level knowledge distillation scheme to guide the detection of marine salient objects. As depicted in Fig. \ref{1-1The_Overall_Framework}, we first design a word-level semantic saliency extraction to adaptively identify salient terms described in the given text through region-word similarity matching. Then, these high-level text features transfer contextual semantic information to the conditional feature learning network based on semantic knowledge distillation, guiding it to generate region-level visual features as object-focused diffusion conditions. To refine the segmentation of fine-grained structures, we develop a dedicated consensus deterministic sampling to suppress inaccurate segmentation caused by overconfidence in camouflaged marine objects.


Our key contributions are summarized as follows:

\begin{itemize}
	\item We propose DiffMSS, a novel object-focused diffusion model for marine saliency segmentation. It simplifies the challenging MSS task into a series of identification, segmentation, and refinement procedures.
	
	\item We design region-level semantic knowledge distillation to capture fine-grained visual features as guiding conditions for object-focused diffusion. We also propose a dedicated CDS scheme to suppress overconfident missegmentations in camouflaged instances. 
	
	
	\item Comprehensive experiments on the public datasets validate that our DiffMSS surpasses existing state-of-the-art solutions in both qualitative and quantitative outcomes.
\end{itemize}

\section{Related Work}\label{Related Work}

\subsection{Marine Saliency Segmenatation}
Existing MSS methods can be roughly divided into handcrafted feature-based methods \cite{Jia2024Visual, Priyadharsini2019USD} and deep learning-based methods \cite{Jin2024USD, Islam2021SVAM, Islam2020SUIM, Hong2023USOD10k, Li2021MarineAS}. Early handcrafted feature-based methods relied on low-level visual features to achieve segmentation \cite{Chen2020Underwater, Peng2024BlurrinessUSD}. With the rise and advancement of visual foundation models, various network architectures have been proposed to address MSS. Li et al. \cite{Li2021MarineAS} proposed a feature interaction encoder and cascaded decoder to extract more comprehensive information, while Liu et al. \cite{Liu2022USD} combined channel and spatial attention modules to refine feature maps to obtain better object boundaries. Although these CNN-based models are effective, they cannot capture the long-range dependencies of complex marine objects and ignore the connectivity between discrete pixels \cite{Liu2024Auto}. Recently, instead of linearly stacking multiple convolutional layers in the network, several deep learning-based USD methods \cite{Chen2023Lightweight, Hong2023USOD10k, Lian2024Diving, Gao2024Pe_Transformer, Deng2023RMFormer} incorporated visual transformers with wider receptive fields into their deep architectures. This way alleviates the computational burden brought by convolution to some extent, but these methods are unreliable in capturing saliency information by improving the encoder architecture.


\subsection{Text-supervised Feature Matching}
With the rise of text-supervised semantic segmentation, many studies \cite{Yi2023TextOS, Wu2024Co_Decomposition, Qu2024VCP-CLIP, Cha2023TCL} have utilized text prompts to enhance segmentation performance. These large vision-language models \cite{Radford2021CLIP, Li2022GLIP} have been trained on large text-image datasets such as LAION-5B \cite{Schuhmann2022Laion}, enabling them to understand the alignment between text descriptions and visual elements. 
They train image and text encoders to align image-text pairs within a joint embedding space, thereby generating segmentation results with zero-shot supervision \cite{Pang2024open, Das2024Mta}. 
Although straightforward, images may contain multiple object instances, and the semantic features of the text should match corresponding segments rather than the entire image. Several region-text alignment methods \cite{Kim2023RegionText, Song2019RegionText} have been proposed to strengthen the consistency between the segmented region and the text description, which enables the network to focus on segmenting the relevant regions described in the text. 

\subsection{Diffusion-based Image Segmentation}
With their powerful generative capabilities, diffusion models have achieved impressive performance in terms of image restoration \cite{Tang2023DM-based}, object detection \cite{Zhao2024Focusdiffuser}, and depth estimation \cite{Zhang2024Atlantis}. By leveraging the adaptive characteristics of the diffusion process, diffusion models have shown potential in various segmentation tasks \cite{Tian2024Segmentation, Pnvr2023LD_ZNet}. For instance, DiffuMask \cite{Wu2023Diffumask} utilizes cross-modal attention maps between image features and conditional text embeddings to segment the most prominent object indicated by text prompts. 
LD-ZNet \cite{Pnvr2023LD_ZNet} performs text-based synthetic image segmentation by revealing rich semantic information within its internal features. However, existing diffusion models employ pixel-level corruption to generate the noised mask directly from the GT, which causes the model to mistakenly assume that the restored contours from the noised mask are accurate \cite{Wang2023Dformer}. In addition, these methods often produce conditional features with limited discriminative representation. To address this, we propose a conditional feature learning network under the guidance of region-level semantic knowledge distillation to robustly generate discriminative conditional features.

\section{Methodology}\label{Methodology}
We first introduce the word-level semantic saliency extraction for identifying words that describe salient objects in Section \ref{Word-level Semantic Saliency Extraction via Word-Region Matching}. Then, Section \ref{Semantic Saliency Knowledge Distillation} presents the semantic knowledge distillation for guiding the conditional feature learning network to generate region-level features as conditions in diffusion. Finally, we describe object-focused conditional diffusion and consensus deterministic sampling for segmenting fine-grained masks in Section \ref{Saliency Diffusion and Sampling}.

\subsection{Word-Level Semantic Saliency Extraction via Word-Region Matching}\label{Word-level Semantic Saliency Extraction via Word-Region Matching}
Unlike image-text alignment, region-word matching focuses on aligning segmented regions (rather than the whole image) with words in a joint embedding space. It ensures consistency between the segmented region and textual description by learning key salient objects in the image.

\textbf{Image-Text Segmenter.} We first introduce an image segmenter and a text segmenter: the former decomposes an image into region segments, while the latter decomposes a text into word segments. It enables both the image and text segmenters to learn region-word consensus when segmenting the input image $\mathcal{I}$ with a paired text $\mathcal{T}$. Specifically, given an image $\mathcal{I}\in \mathbb{R}^{H\times W\times C_{v}}$ and the corresponding text $\mathcal{T}\in \mathbb{R}^{N_{t}\times C_{t}}$, where $H,  W, C_{v}$ represent the height, width, channel of image $\mathcal{I}$, and $N_{t}, C_{t}$ represent the number, dimension of the words. We utilize the image segmenter and text segmenter to process the image-text pairs, thus obtaining a group of $M$ region masks $\mathbb{X}^{v}=\left\{\mathcal{X}_{i}^{v} \right\}_{i=1}^{M}$ and the corresponding text $\mathbb{X}^{t}=\left\{\mathcal{X}_{j}^{t} \right\}_{j=1}^{N}$ of $N$ single-word nouns. That is, $\mathbb{X}^{v}$ contains several sub-images $\mathcal{X}_{i}^{v}$ obtained by cropping and masking relevant regions from the input image $\mathcal{I}$, while $\mathbb{X}^{t}$ takes a text $\mathcal{T}$ of length $N$ as input and extracts each word $\mathcal{X}_{j}^{t}$ in $\mathcal{T}$.
\begin{figure}[!tp]
	\setlength{\abovecaptionskip}{0.1cm}
	\setlength{\belowcaptionskip}{-0.2cm}
	\centering
	\includegraphics[width=0.48\textwidth]{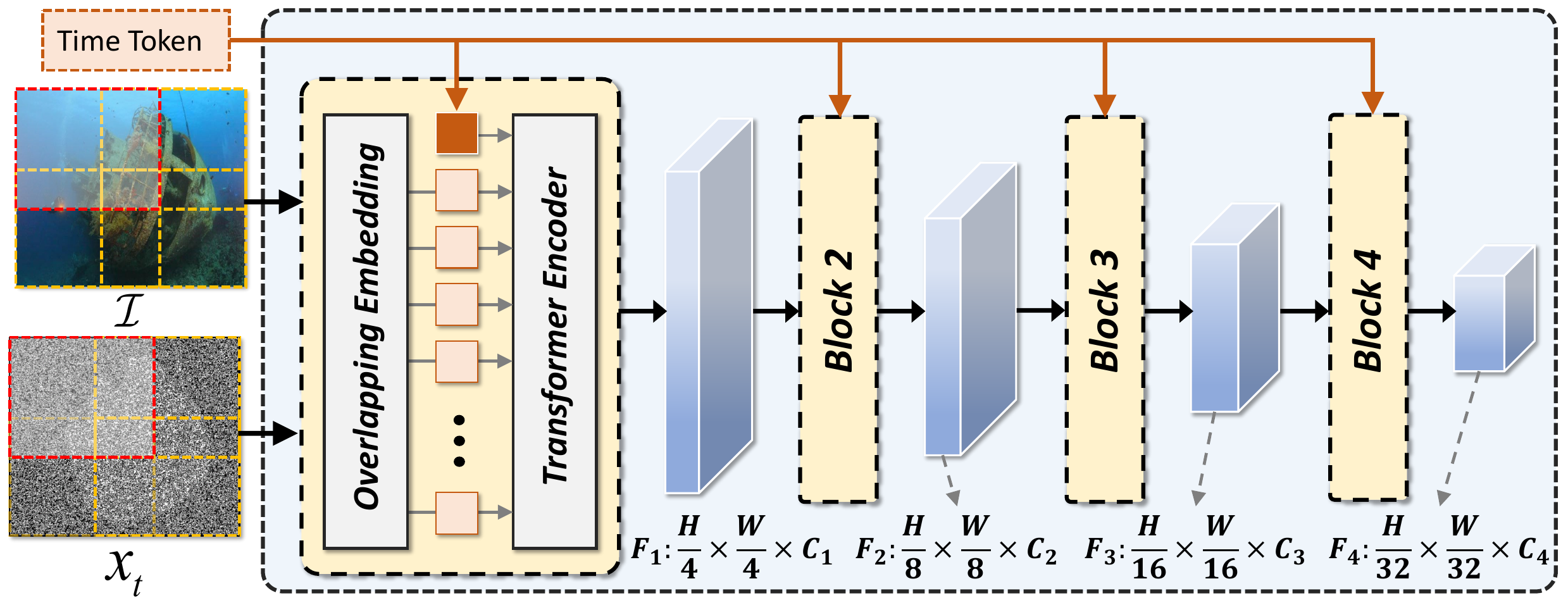}
	\caption{Illustration of the Conditional Features Learning Network (CFLN). It cascades the input image $\mathcal{I}$, intermediate sample $\mathbf{x}_{t}$, and time token $t$ through four Transformer-based blocks to extract region-level features as conditions in diffusion.}
	\label{1-2Conditional_Features_Learning_Network}
    \vspace{-0.1cm}
\end{figure}

\textbf{Saliency Word-Token Discover}. We then select high-confidence salient words as the guided tokens, instead of the whole caption. We employ the image encoder $E_{v}(\mathcal{X}_{i}^{v})$ and text encoder $E_{t}(\mathcal{X}_{j}^{t})$ of the pre-trained CLIP model \cite{Radford2021CLIP} to extract semantic saliency features from the highlighted regions and words, respectively. For each input image, the visual embedding tokens $\mathcal{F}_{i}^{v}$ are calculated as follows:

\begin{equation}\label{visual_embedding_tokens}
\mathcal{F}_{i}^{v}=\mathcal{W}^{v}\times \mathcal{Z}^{v}_{i}, i \in \left\{1, 2, \dots, M \right\}, 
\end{equation}
where $\mathcal{Z}^{v}_{i}$ represents the visual features provided by the image encoder $\mathcal{Z}^{v}_{i}=E_{v}(\mathcal{X}_{i}^{v})$, and $\mathcal{W}^{v}$ is the projection matrix that converts $\mathcal{Z}^{v}_{i}$ into the vision embedding tokens $\mathcal{F}_{i}^{v}$. In the same way, the word prompts $\mathcal{X}_{j}^{t}$ are transformed into textual embedding tokens $\mathcal{F}_{j}^{t}$ through the projection matrix $\mathcal{W}^{t}$. Both types of tokens have the same dimensionality in the joint embedding space and the region-word similarity is calculated as follows:
\begin{equation}\label{Scores_Calculation}
R_{k}=\frac{1}{MN}\sum_{i=1}^{M}\sum_{j=1}^{N} Softmax\left(\mathcal{F}_{i}^{v}{\mathcal{F}_{j}^{t}}^{T}\right).
\end{equation}

After that, we calculate the average $m =\text{Mean}(R_{k})$ of obtained similarity scores and then select the indices of scorers exceeding $m$ from the candidate list containing $L_{t}$ tokens with priority. Mathematically, it is expressed as:
\begin{equation}\label{Raters_selection}
\mathcal{F}^{st} = \left\{k \mid R\{k\} \ge m, \, k \in \left\{1, 2, \dots, L_{t} \right\} \right\},
\end{equation}
where $\mathcal{F}^{st}$ defines the index of selected tokens from the candidate list with the score priority. To avoid region-word mismatches, we use the noun-selector \cite{Devlin2018Bert} to filter the segmented words. That is, some words that are irrelevant to the salient objects in the visual domain, such as prepositions and pronouns, are not considered for guiding the conditional feature generation.


\subsection{Region-Level Semantic Saliency Knowledge Distillation}\label{Semantic Saliency Knowledge Distillation}
By leveraging high-level semantic tokens and aligning them with visual features, it can guide the diffusion model to focus on the salient regions, thereby improving the accuracy of object segmentation.


\textbf{Conditional Features Learning Network.} The network aims to generate region-level conditions that enable the diffusion model to effectively identify salient objects at each denoising step. Its design needs to meet three requirements: 1) Extracting discriminative salient features based on the image content; 2) Providing region-level conditions associated with the current denoising step; and 3) Capturing long-range dependencies and contextual information of the whole image. In addition, the inherent degradation characteristics of underwater images significantly interfere with the extraction of discriminative image features, thus diminishing the performance of the mask decoder.

To address this issue, we design a well-generalized Conditional Feature Learning Network (CFLN) based on the pyramid vision transformer \cite{Wang2022PVT}. As shown in Fig. \ref{1-2Conditional_Features_Learning_Network}, it extracts visual features $\{\mathcal{F}_l\}_{l=1}^4$ from a triplet data $(\mathcal{I}, \mathbf{x}_t, t)$, in which $\mathcal{I}$ represents the input image, $\mathbf{x}_{t}$ denotes previous sampling results, and $t$ represents the denoising step. $\mathbf{x}_t$ serves as a guiding cue to assist CFLN in adaptively focusing on specific regions, while adding the time step $t$ aims to improve the synchronization of the extracted conditions in the denoising step. To achieve this, we employ the zero overlap embedding to incorporate the noise mask $\mathbf{x}_t$ into the first block in a controlled manner without disrupting the Transformer structure, which is expressed as follows:
\begin{equation}\label{Zero_Overlap_Embedding}
\mathcal{F}_{l}= 
\begin{cases}
	Norm\left(\mathbb{R}\left({Conv}(\mathcal{I}) + {Conv}_z(\mathbf{x}_t)\right)\right),& l = 1,\\
	Norm\left(\mathbb{R}({Conv}(\mathcal{F}_{l-1}))\right),& l \neq 1.
\end{cases}
\end{equation}
where ${Conv}(\mathcal{I})$ and ${Conv}_z(\mathbf{x}_t)$ denote convolutional layers, differing in whether the weights and biases are initialized to zero. $Norm(\cdot)$ denotes layer normalization, while $\mathbb{R}(\cdot)$ represents transforming the feature map into tokens.

In addition to embedding the noise mask, we desire that the CFLN can adaptively tune the conditional features over time steps. We propose a scheme to concatenate the time token $t$ with the embedding patches $\mathcal{F}_{l}$, as follows:
\begin{equation}\label{Time_Token_Concatenation}
\mathcal{F}_l^{v} = \mathbb{R}^{-1}\left(MHA([t; \mathcal{F}_{l}])\right), l \in \left \{1,2,3,4 \right\},
\end{equation}
where $[;]$ refers to the connection operation, $\mathbb{R}^{-1}$ reconverts tokens into multi-scale features, and $MHA$ represents the multi-head attention.

\textbf{Word-level Knowledge Transfer.}
In Section \ref{Word-level Semantic Saliency Extraction via Word-Region Matching}, we have obtained word-level tokens that contain semantic information, which assists in identifying salient objects within the entire image. Based on this, we utilize the Semantic Knowledge Distillation (SKD) to constrain the generation of diffusive conditions throughout the training phase. Specifically, we design two distinct projectors to map the textual features of tokens (denoted as $\mathcal{F}^{st}$) and the visual features of conditions (denoted as $\mathcal{F}_l^{v}$) into a unified latent feature space. In other words, $\mathcal{F}^{st}$ are selected word-level tokens derived from the text encoder, while $\mathcal{F}_l^{v}$ represents the conditional features generated by the CFLN module. Considering that these two features should exhibit consistency across the latent space, we define a consistent loss $\mathcal{L}_{\textit{consist}}$ to constrain them, expressed as:
\begin{equation}\label{Consistency_Constraint_Loss}
	\mathcal{L}_{\textit{consist}}= - \frac{1}{N} \sum_{i=1}^{N} \frac{{Proj}(\mathcal{F}_l^{v}(i))_v \cdot {Proj}(\mathcal{F}^{st}(i))_t}{\|{Proj}(\mathcal{F}_l^{v}(i))_v\|_2 \, \|{Proj}(\mathcal{F}^{st}(i))_t\|_2},
\end{equation}
where ${Proj}(\cdot)_t$ and ${Proj}(\cdot)_v$ represent the projectors for mapping textual tokens and conditional features into the latent embedding space, respectively.

For the discrete features, we employ the Local Emphasis (LE) module in \cite{Wang2022LE} and convolutional calculation to aggregate them, expressed as follows:
\begin{equation}\label{Feature_Aggregation_Loss}
\mathcal{F}_l^{v} = Conv([\mathcal{F}_{l+1}^{v}, LE(\mathcal{F}_l^{v})]), l \in \{3, 2, 1\},
\end{equation}
where the aggregated feature is defined as $\mathcal{F}_a^{v}=LE(\mathcal{F}_4^{v})$ and serves as the region-level diffusion conditions.

\subsection{Object-Focused Diffusion and Sampling}\label{Saliency Diffusion and Sampling}
Compared to traditional segmentation baselines, our proposed DiffMSS framework employs a revised conditional diffusion model with feature consistency learning to generate predicted masks. However, iterative diffusion and sampling may face two inherent challenges when generating masks: 1) Restoring a high-fidelity mask from low signal-to-noise ratio noise based on visual features is challenging; 2) Degraded images may cause well-trained models to produce occasional missegmentations due to overconfidence. The reason for this dilemma is that the model tends to choose the path of least resistance for parameter learning. They rely on more obvious noise masks instead of utilizing conditional features for generation. To address these issues, we propose the Object-Focused Conditional Diffusion (OFCD) and Consensus Deterministic Sampling (CDS) to achieve fine-grained structural segmentation for marine camouflage objects.

\textbf{Object-Focused Conditional Diffusion.} In forward diffusion, given a training sample $x_{0}\sim q(x_{0})$, the noised samples $\{x_{t}\}_{t=1}^T$ are obtained according to the following Markov process:
\begin{equation}\label{Forward_process_function}
q(\mathrm{x}_{t}|\mathrm{x}_{t-1})=\mathcal N\left(\mathrm{x}_{t};\sqrt{1-\beta_{t}}\mathrm{x}_{t-1},\beta_{t}\mathbf{I}\right),
\end{equation}
where $\beta_{t}$ denote the pre-defined noise
schedule at $t$-th time step. The marginal distribution of $x_{t}$ can be described as:
\begin{equation}\label{Forward_fromX0_function}
q(\mathrm{x}_{1:T}|\mathrm{x}_{0})=\mathcal N\left(\mathrm{x}_{t};\sqrt{\bar{\alpha}_{t}}\mathrm{x}_{0},(1-\bar{\alpha}_{t})\mathbf{I}\right),
\end{equation}
where $\alpha_{t}=1-\beta_{t}$ and $\bar{\alpha}_{t}={\textstyle\prod_{i=1}^{t}\alpha_{i}}$.
\begin{figure}[!tp]
	\setlength{\abovecaptionskip}{0.1cm}
	\setlength{\belowcaptionskip}{-0.2cm}
	\centering
	\includegraphics[width=0.47\textwidth]{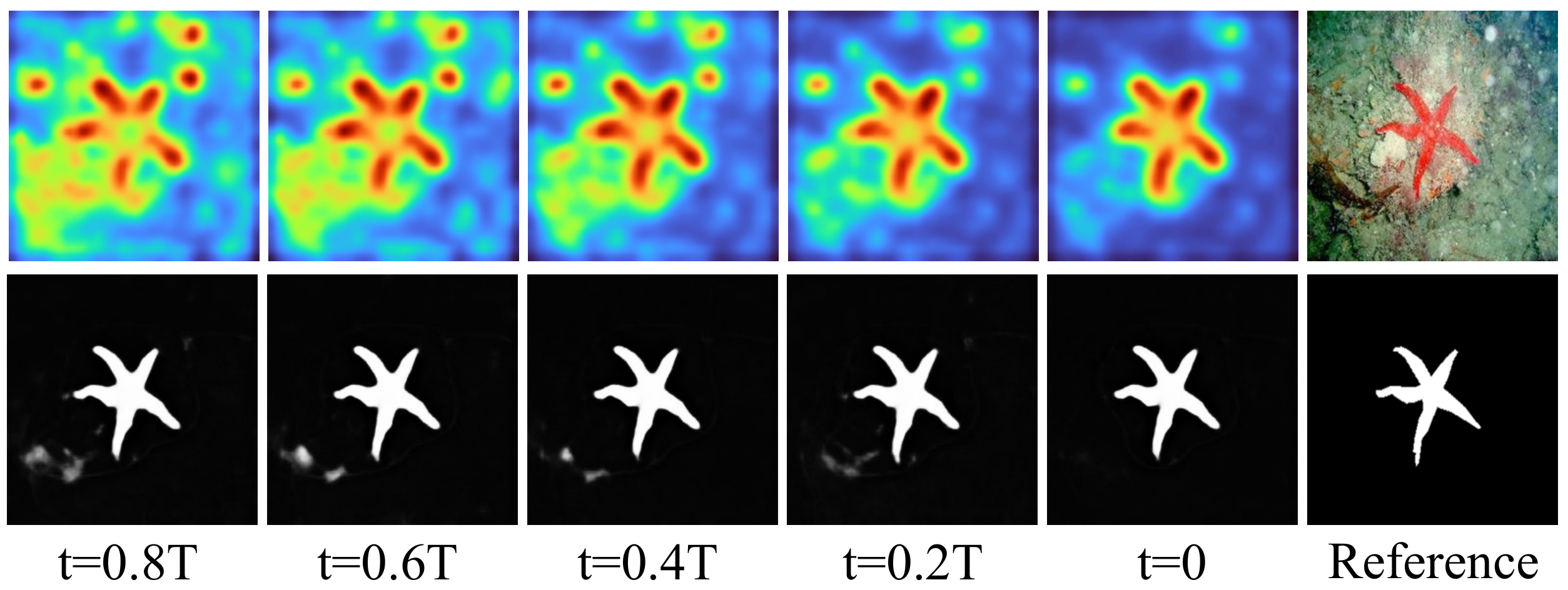}
	\caption{The conditional feature maps and mask predictions at different sampling steps $t$. }
	\label{1-3Mask_and_Feature}
	\vspace{-0.2cm}
\end{figure}

The previous diffusion paradigm that learns a conditional reverse process $p_{\theta}(\mathrm{x}_{T:0}|\mathrm{y})$ without modifying the forward diffusion $q(\mathrm{x}_{1:T}|\mathrm{x}_{0})$, ensuring the sampled $\hat{\mathrm{x}}_{0}$ is faithful to the raw data distribution. Instead of taking input image $\mathrm{y}$ as an invariant condition, we employ the aggregated features $\mathcal{F}_{a}^{v}$ produced by the CFLN module as conditions. Mathematically, it is expressed as follows:

\begin{equation}\label{Revised_Inverse_process_function}
q(\mathrm{x_{t-1}|x}_{t},\mathcal{F}_{a}^{v})=\mathcal N\left(\mathrm{x}_{t-1};\mu_{\theta}(\mathrm{x}_{t},\mathcal{F}_{a}^{v},t),\delta^{2}_{t}\mathbf{I}\right),
\end{equation}
where the variance $\delta^{2}_{t}=\frac{1-\bar{\alpha}_{t-1}}{1-\bar{\alpha}_{t}}\beta_{t}$, and the mean $\mu_\theta$ is defined as follows:
\begin{align}\label{Predicted_mean_function}
\mu_{\theta}(\mathrm{x}_{t}, \mathcal{F}_{a}^{v},  t)=\frac{1}{\sqrt{\bar{\alpha}_{t}}}(\mathrm{x}_{t}-\frac{\beta_{t}}{\sqrt{1-\bar{\alpha}_{t}}}\epsilon_{\theta}(\mathrm{x}_{t}, \mathcal{F}_{a}^{v}, t)),
\end{align}
where $\epsilon_{\theta}(\mathrm{x}_{t}, \mathcal{F}_{a}^{v}, t)$ represents the predicted noise by optimizing the parameters $\theta$ of our proposed DiffMSS model. We transform the estimated noise $\epsilon_{\theta}$ into the salient mask conditioned on the region-level aggregated features $\mathcal{F}_{a}^{v}$, which is defined as follows:
\begin{align}\label{Predicted_result_function}
\hat{\mathrm{x}}_0 =\frac{\mathrm{x}_t-\sqrt{1-\bar{\alpha}_t} \, \epsilon_{\theta}(\mathrm{x}_{t},\mathcal{F}_{a}^{v}, t)}{\sqrt{\bar{\alpha}_t}},
\end{align}
where $\hat{\mathrm{x}}_0$ is predicted by our model $f_\theta(\mathrm{x}_{t}, \mathcal{F}_{a}^{v}, t)$. Based on this, we utilize the saliency mask $\mathrm{x}_0$ corresponding to the real-world underwater scene as a reference to constrain the rationality of the predicted mask, as expressed below:
\begin{align}\label{Mask_Loss_function}
\mathcal{L}_{\textit{mask}}= \mathcal{L}_{\textit{BCE}}^{w}(\hat{\mathrm{x}}_0,\mathrm{x}_0)+\mathcal{L}_{\textit{IoU}}^{w}(\hat{\mathrm{x}}_0,\mathrm{x}_0).
\end{align}

Based on the semantic knowledge distillation term $\mathcal{L}_{consist}$ and saliency mask refinement term $\mathcal{L}_{mask}$, the hybrid objective function $\mathcal{L}_{total}$ is defined by combining them as follows:
\begin{equation}\label{Total_loss_function}
\mathcal{L}_{Total}=\mathcal{L}_{consist}+\lambda\mathcal{L}_{mask},
\end{equation}
where $\lambda=0.5$ is weighted to coordinate the significance of each term in the experiment.



\begin{figure*}[t]
	\setlength{\abovecaptionskip}{0.1cm}
	\setlength{\belowcaptionskip}{-0.2cm}
	\centering
	\includegraphics[width=1.0\textwidth]{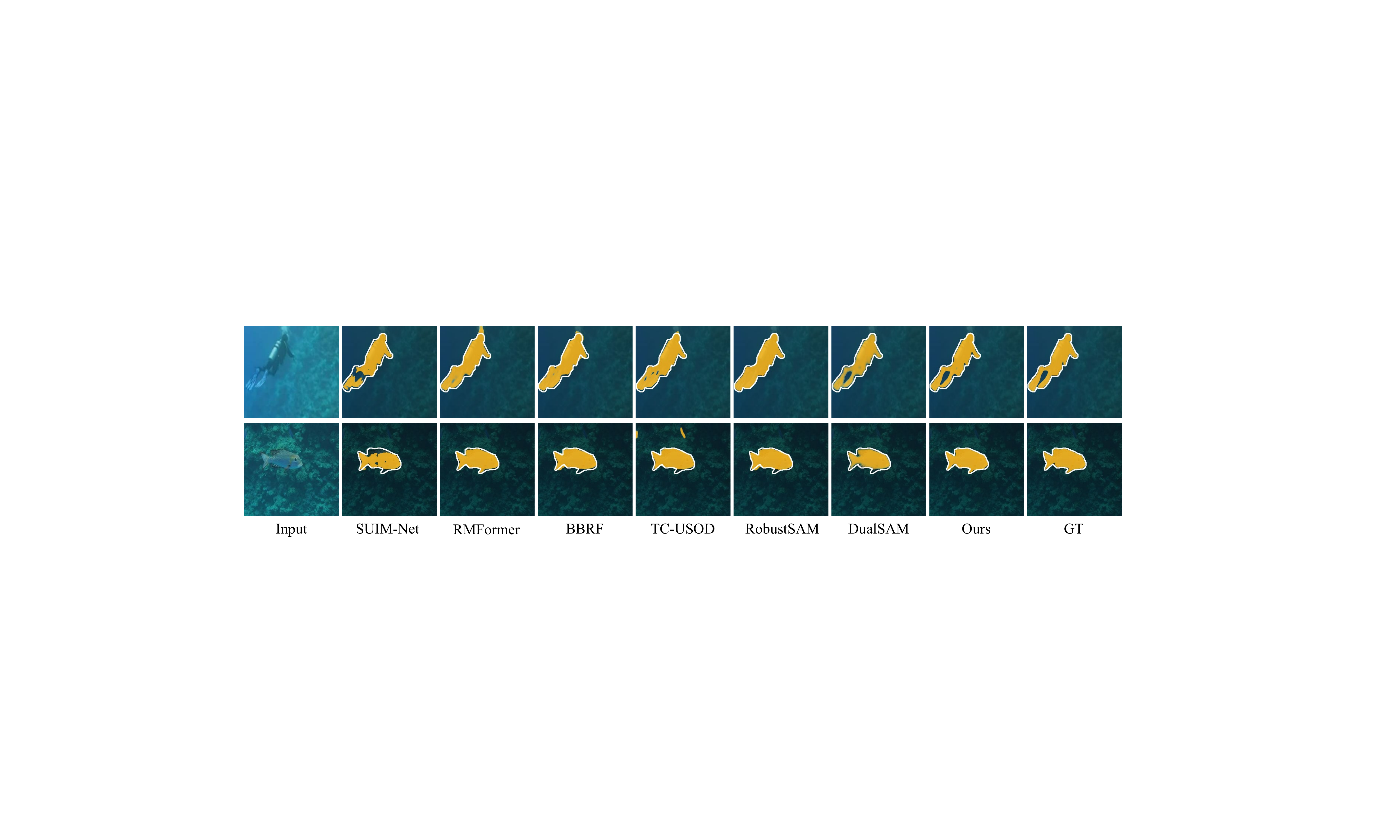}
	\caption	
	{Visualization comparisons between our DiffMSS and state-of-the-art methods on the common underwater salient objects. The segmentation results are marked in orange.}
	\label{4-1Qualitative_Evaluation_of_First_Saliency_Detection}
	\vspace{-0.1cm}
\end{figure*}
\begin{figure*}[h]
	\setlength{\abovecaptionskip}{0.1cm}
	\setlength{\belowcaptionskip}{-0.2cm}
	\centering
	\includegraphics[width=1.0\textwidth]{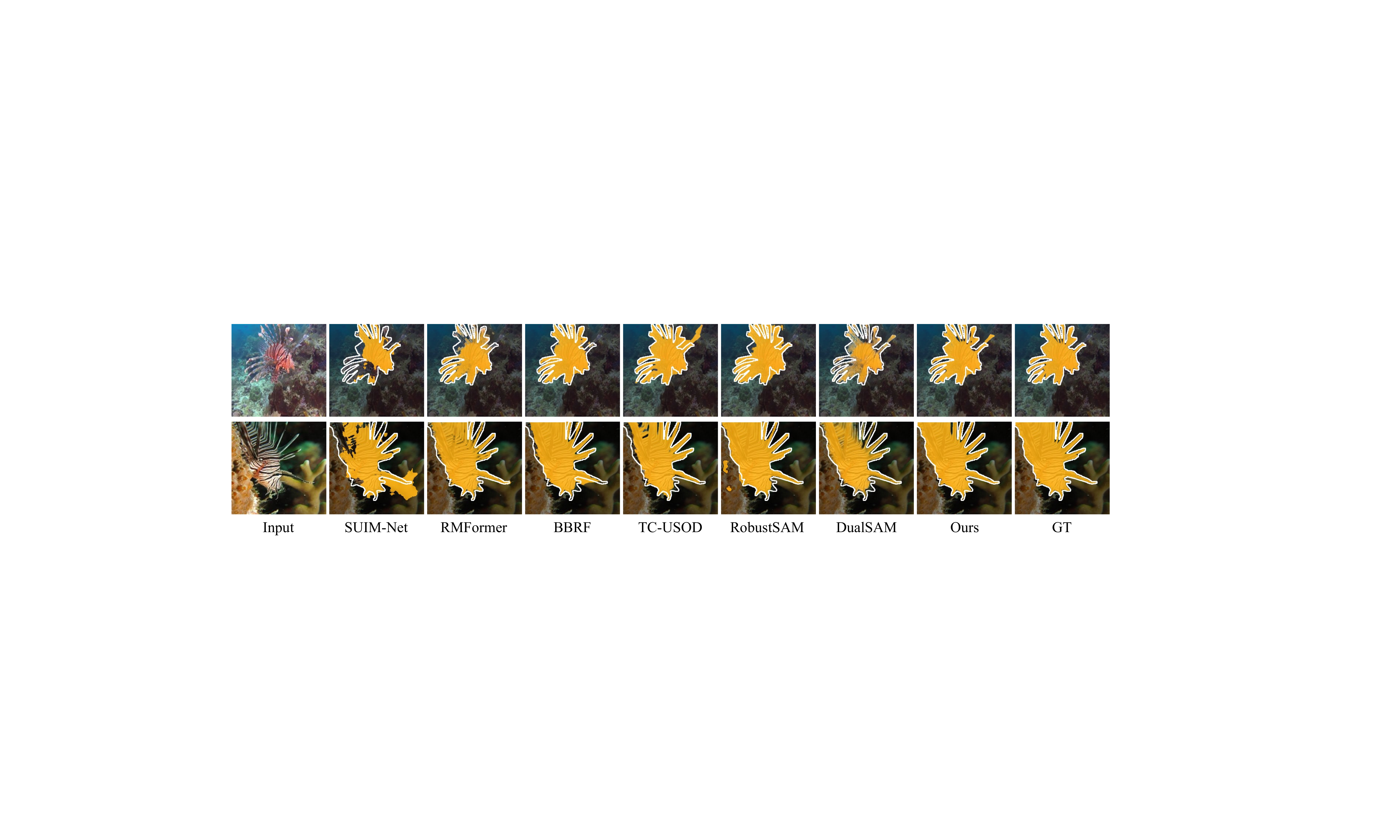}
	\caption	
	{Visualization comparisons between our DiffMSS and state-of-the-art methods on the challenging underwater camouflage objects with fine-grained structures. The segmentation results are marked in orange.}
	\label{4-2Qualitative_Evaluation_of_Second_Saliency_Detection}
	\vspace{-0.2cm}
\end{figure*}

To illustrate that DiffMSS can reduce noise and progressively focus on salient objects, we display the predicted results and conditional feature maps captured at different sampling steps in Fig. \ref{1-3Mask_and_Feature}. It is clear that the model progressively focuses on salient objects and refines the mask, enabling it to establish well-defined boundaries based on the foreground objects.

\textbf{Consensus Deterministic Sampling.}
To improve the segmentation accuracy of fine-grained anatomical structures in marine instances, we introduce a Consensus Deterministic Sampling (CDS) method to aggregate predictions from each denoising step, which is inspired by the saliency detection annotation in \cite{Zhang2021UncertaintySD}. Specifically, we denote the denoised image $\hat{\mathrm{x}}_0$ as $P_t$ at each sampling stage $t$. After obtaining multiple predictions $\{P_t\}_{t=1}^T$, which are then calculated as binary masks by setting an average threshold. These predictions $\{P_t^b\}_{t=1}^T$ vote on the position of each point to generate a candidate mask. The probability value of each selected point is calculated as the average of all predictions. Mathematically, it is defined as follows:
\begin{equation}\label{noise_loss_function}
M_{pre}= \left\lfloor \frac{1}{T}\sum_{t=1}^{T} P_t^b + \varphi \right \rfloor  \ast Norm\left(\frac{1}{T}\sum_{t=1}^{T} P_t\right),
\end{equation}
where $\varphi=0.5$ represents the average threshold calculated with samples. The CDS schedule generates multiple predictions by iterative sampling, which enables us to improve mask accuracy through ensemble techniques.

\textbf{Training and Inference.} The training phase of DiffMSS requires the degraded image $\mathcal{I}$, the corresponding caption $\mathcal{T}$, and the reference saliency mask $\mathrm{x}_{0}$ for supervision, whereas its inference only requires the degraded image $\mathcal{I}$ as input. In other words, the inference of DiffMSS relies solely on Object-Focused Conditional Diffusion (OFCD) to generate segmentation results, without the Word-level Semantic Saliency Extraction (WSSE) procedure. More detailed training and sampling procedures can be found in the supplementary material.

\section{Experiments}\label{Experiments}

\subsection{Experimental Setups}
\textbf{Implementation Details.} The proposed DiffMSS is trained using the Pytorch framework on two NVIDIA GeForce RTX 4090 GPUs for $150$ epochs. During the training phase, the batch size and patch size are set to 32 and $256\times256$, respectively. The Adam optimizer comes with an initial learning rate of $1\times10^{-4}$ and decreases it by a factor of 0.8 after every ten
epochs. The diffusion steps are set to $T=1000$ with a noise schedule $\beta_{t}$ that increases linearly from 0.0001 to 0.02, while the sampling steps are set to $S=10$ for efficient restoration. More detailed hyperparameter settings can be found in the supplementary material.

\begin{table*}[!htp]\small
    \setlength{\abovecaptionskip}{0.1cm}
    \renewcommand\arraystretch{1.0}
    \tabcolsep 0.076 in
    \centering
    \caption{Quantitative evaluation of our DiffMSS and state-of-the-art methods on three public underwater datasets (\textit{USOD10K} \cite{Hong2023USOD10k}, \textit{SUIM} \cite{Islam2020SUIM}, and \textit{UFO-120} \cite{Islam2020UFO120}). The best and second-best results are highlighted with \textbf{bold} and \underline{underlined}, respectively.}
    \resizebox{\linewidth}{!}{
        \begin{tabular}{lcccc|cccc|cccc}
            \hline
            \multirow{2}{*}{Config.}&\multicolumn{4}{c|}{\textbf{USOD10K}}&\multicolumn{4}{c|}{\textbf{SUIM}}&\multicolumn{4}{c}{\textbf{UFO-120}}\\
            \cline{2-13}
            &$F_{\beta}^{w}\uparrow$ &$E_{\phi}^{m}\uparrow$ & $S_{\alpha}\uparrow$ & $M_{AE}$ $\downarrow$  &$F_{\beta}^{w}\uparrow$ &$E_{\phi}^{m}\uparrow$ & $S_{\alpha}\uparrow$ & $M_{AE}$$\downarrow$ &$F_{\beta}^{w}\uparrow$ &$E_{\phi}^{m}\uparrow$ & $S_{\alpha}\uparrow$ & $M_{AE}$ $\downarrow$\\
            \hline
            \text{SUIM-Net} \cite{Islam2020SUIM} &0.783 &0.856 &0.797 &0.1011 &0.807 &0.867 &0.826 &0.0787 &0.734 &0.751 &0.739 	&0.1162 \\
            \text{RMFormer} \cite{Deng2023RMFormer} &0.828 &0.910 &0.867 &0.0439 &0.830 &0.908 &0.859 &0.0623 &0.829 &0.865 &0.817 &0.0942 \\
            \text{BBRF} \cite{Ma2023BBRF} &0.902 &0.935 &0.913 &0.0317 &0.856 &0.891 &0.856 &0.0679 &0.847 &0.876 &0.839 &0.0695 \\
            \text{TC-USOD} \cite{Hong2023USOD10k} &\underline{0.910}  &0.953 &0.912	&0.0236 &\underline{0.879} &\textbf{0.951} &\underline{0.893} &\underline{0.0388} &0.856 &0.917 &0.859 &\underline{0.0631}\\
            \text{RobustSAM} \cite{Chen2024RobustSAM} &0.897 &0.946 &0.909 &0.0356 &0.861 &0.924 &0.869 &0.0567 &0.839 &0.893 &0.847 &0.0717\\
            \text{DualSAM} \cite{Zhang2024DualSAM} &0.909 &\textbf{0.959} &\underline{0.916} &\underline{0.0218} &0.876 &0.937 &0.881 &0.0465 &\underline{0.858} &\underline{0.921} &\underline{0.861} &0.0637\\
            \text{DiffMSS} &\textbf{0.912} &\underline{0.956} &\textbf{0.922} &\textbf{0.0203} &\textbf{0.891} &\underline{0.947} &\textbf{0.908} &\textbf{0.0376} &\textbf{0.867} &\textbf{0.927} &\textbf{0.873} &\textbf{0.0566}\\
            \hline
    \end{tabular}}
    \label{Saliency_Segmentation_Metrics}
    \vspace{-0.2cm}
\end{table*}

\textbf{Benchmark Datasets.} We evaluate DiffMSS on three popular USOD benchmarks (USOD10K \cite{Hong2023USOD10k}, SUIM \cite{Islam2020SUIM} and UFO-120 \cite{Islam2020UFO120}), all of which are real-world underwater images with references. Specifically, we follow the default settings in USOD10K, using 7,178 images for training and 1,026 images for testing. Meanwhile, we use 1,300 images from each of the SUIM and UFO-120 datasets for training, with the remaining images reserved for testing, respectively. For a fair comparison, all compared methods are retrained on the same data with their default settings.

\textbf{Evaluation Metrics.} We adopt five commonly used metrics for MSS tasks evaluation, including weighted F-measure ($F_{\beta}^{w}$) \cite{Margolin2014F-measure}, max E-measure ($E_{\phi}^{m}$) \cite{Fan2018E-measure}, S-measure ($S_{\alpha}$) \cite{Fan2017S-measure}, and mean absolute error ($M_{AE}$) \cite{Perazzi2012MAE}. 

\subsection{Comparison with State-of-the-Arts}
We compare the proposed DiffMSS model with six state-of-the-art (SOTA) saliency object detection methods, including SUIM-Net \cite{Islam2020SUIM}, RMFormer \cite{Deng2023RMFormer}, BBRF \cite{Ma2023BBRF}, TC-USOD \cite{Hong2023USOD10k}, RobustSAM \cite{Chen2024RobustSAM}, and DualSAM \cite{Zhang2024DualSAM}.

\textbf{Qualitative Evaluation.} As shown in Fig. \ref{4-1Qualitative_Evaluation_of_First_Saliency_Detection}, we first conduct a visual comparison of our DiffMSS with several SOTA methods on two common underwater salient objects. Compared with SUIM-Net, RobustSAM, and DualSAM, our model shows superior segmentation performance, especially involving objects with blurred boundaries. We then evaluate DiffMSS on the challenging marine camouflage objects with fine-grained structures. As shown in Fig. \ref{4-2Qualitative_Evaluation_of_Second_Saliency_Detection}, our model consistently achieves superior segmentation results, characterized by well-defined boundaries and strong robustness against underwater noise and artifacts.

\textbf{Quantitative Evaluation.} We further conduct a quantitative evaluation of these compared methods, and the results are presented in Table \ref{Saliency_Segmentation_Metrics}. DiffMSS consistently achieves the best or second-best scores across all metrics on the three public underwater datasets, especially performing well on UFO-120. This demonstrates the robustness and generalization ability of our model in handling marine saliency segmentation tasks under various challenging conditions.

\subsection{Evaluation of Model Efficiency}
\textbf{Parameters and FLOPs.}	Considering the limited computational resources of underwater embedded devices, our DiffMSS ensures segmentation accuracy while excelling in terms of parameters and FLOPs. As shown in Table \ref{Parameters_FLOPs_Time}, DiffMSS’s 68.41M parameters are lower than RMFormer (174.19M) and RobustSAM (407.76M). Moreover, our DiffMSS achieves the lowest FLOPs (24.98G) that outperforms other methods like SUIM-Net and TC-USOD, making it a more efficient choice for underwater applications with limited computing resources.
\begin{table}[!htp]\small
	\setlength{\abovecaptionskip}{0.1cm}
	\tabcolsep 0.06 in
	\centering
    \caption{Efficiency of each method with Parameters (M), FLOPs (G), Inference Time (s), and Avg$M_{AE}$. The best and second-best scores are highlighted with \textbf{bold} and \underline{underlined}, respectively.}
	\begin{tabular}{l|cc|cc}
		\hline
		\text{Method} &\text{Param. $\downarrow$} &\text{FLOPs $\downarrow$} &\text{Time $\downarrow$} &\text{Avg$M_{AE}$ $\downarrow$}\\
		\hline
		\text{SUIM-Net} \cite{Islam2020SUIM} &\textbf{12.22} 	&71.46  &0.265 	&0.1006 \\	
        \text{RMFormer} \cite{Deng2023RMFormer} &174.19 &563.14  &0.315 &0.0503 \\
        \text{BBRF} \cite{Ma2023BBRF} &74.01 	&31.13  &0.107 	&0.0385 \\
        \text{TC-USOD} \cite{Hong2023USOD10k} &117.64 &\underline{29.64}  &0.089 	&0.0287 \\
        \text{RobustSAM} \cite{Chen2024RobustSAM} &407.76	&1492.60  &0.214 &0.0409 \\
        \text{DualSAM} \cite{Zhang2024DualSAM} &159.95 &325.68  &\underline{0.088} &\underline{0.0280}\\
        \text{DiffMSS (Ours)} &\underline{68.41} 	&\textbf{24.98}  &\textbf{0.033} &\textbf{0.0253} \\
		\hline
	\end{tabular}
	\label{Parameters_FLOPs_Time}
	\vspace{-0.3cm}
\end{table}


\textbf{Inference Time.} Unlike these compared methods that stack multiple convolutional sequences or Segment Anything Model (SAM)-based, our DiffMSS exploits object-focused conditional diffusion to optimize computational efficiency while maintaining effective deep feature extraction. As shown in Table \ref{Parameters_FLOPs_Time}, DiffMSS achieves the fastest inference time of 0.033s. Although DualSAM's inference time is relatively short (0.081s), it is still more than twice ours. The efficiency is mainly attributed to the semantic knowledge distillation that transfers high-level text semantic information, and the inference requires ten sampling steps to generate predicted results from a single input image.


\subsection{Ablation Study}
\textbf{Ablation Study of Semantic Knowledge Distillation.} We conduct an ablation study with and without semantic knowledge distillation (``-w/ SKD'' and ``-w/o SKD'') and evaluate segmentation performance using image-text matching (``I-T'') or region-word matching (``R-W'') schemes in the ``-w/ SKD'' case. Table \ref{Ablation_study_of_SKD} shows that utilizing ``I-T'' matching can improve the performance of saliency segmentation to a certain extent, but our ``R-W'' matching scheme achieves the highest scores across all metrics, which demonstrates the effectiveness of word-level semantic alignment in achieving object-focused diffusion.
\begin{table}[!htp]\small
	\setlength{\abovecaptionskip}{0.1cm}
	\tabcolsep 0.08 in
	\centering
    \caption{Ablation study of Semantic Knowledge Distillation.}
	\begin{tabular}{cc|cccc}
		\hline
		-w/o SKD &-w/ SKD &$F_{\beta}^{w}\uparrow$ &$E_{\phi}^{m}\uparrow$ & $S_{\alpha}\uparrow$ & $M_{AE}$ $\downarrow$\\
		\hline
		$\checkmark$&  &0.897  &0.942 	&0.913 	&0.0355 \\
		&$\checkmark$ &\textbf{0.912}  &\textbf{0.956}  &\textbf{0.922}  &\textbf{0.0203} \\
		\hline
	\end{tabular}
	\label{Ablation_study_of_SKD}
	\vspace{-0.3cm}
\end{table}
\begin{table}[!htp]\small
	\setlength{\abovecaptionskip}{0.1cm}
	\tabcolsep 0.066 in
	\centering
    \caption{Ablation study of different modality matching schemes for semantic knowledge distillation. ``I-T'' represents image-text matching, while ``R-W'' represents region-word matching.}
	\begin{tabular}{cc|cccc}
		\hline
		I-T Match. &R-W Match. &$F_{\beta}^{w}\uparrow$ &$E_{\phi}^{m}\uparrow$ & $S_{\alpha}\uparrow$ & $M_{AE}$ $\downarrow$\\
		\hline
		$\checkmark$ & &0.887  &0.943  &0.917  &0.0789 \\
		&$\checkmark$ &\textbf{0.912}  &\textbf{0.956}  &\textbf{0.922}  &\textbf{0.0203} \\
		\hline
	\end{tabular}
	\label{Ablation_study_of_semantic_knowledge_distillation}
	\vspace{-0.2cm}
\end{table}

\textbf{Ablation Study of Different Modality Matching.}
In the semantic saliency extraction procedure, we further conduct an ablation study on different modality matching in the ``-w/ SKD'' case, including Image-Text matching (denoted as ``I-T'') or Region-Word matching (denoted as ``R-W''). As shown in Table \ref{Ablation_study_of_semantic_knowledge_distillation}, compared with “-w/o SKD”, ``I-T'' matching can improve the performance of saliency segmentation to a certain extent, but our ``R-W'' matching scheme achieves the highest scores across all metrics, which demonstrates the effectiveness of word-level semantic alignment in achieving object-focused diffusion.

\textbf{Necessity of Features Aggregation in CFLN.} Table \ref{Ablation_study_of_CFLN} presents a discussion on the impact of aggregating different layer features ($\mathcal{F}_1$, $\mathcal{F}_2$, $\mathcal{F}_3$, $\mathcal{F}_4$) as diffusion conditions in the CFLN module. The scores of the four metrics gradually increase with the aggregation of more feature layers. All features are aggregated together to produce the highest $F_{\beta}^{w}$, $E_{\phi}^{m}$, and $S_{\alpha}$, along with the lowest $M_{AE}$, while the number of model parameters and computational burden increase only slightly compared to the former.
\begin{table}[!htp]\small
	\setlength{\abovecaptionskip}{0.1cm}
	\renewcommand\arraystretch{1.0}
	\tabcolsep 0.034 in
	\centering
    \caption{Ablation study of aggregating different layer features as region-level diffusion conditions in CFLN module.}
	\begin{tabular}{cccc|cccc|cc}
		\hline
		$\mathcal{F}_1$ &$\mathcal{F}_2$ &$\mathcal{F}_3$ &$\mathcal{F}_4$ &$F_{\beta}^{w}\uparrow$ &$E_{\phi}^{m}\uparrow$ & $S_{\alpha}\uparrow$ & $M_{AE}$ $\downarrow$ & Param. & FLOPs\\
		\hline
		$\checkmark$ & & & &0.664 &0.853 &0.796 	&0.1219  &65.23 &19.49\\
        $\checkmark$ &$\checkmark$ & & &0.831 	&0.943 	&0.879 	&0.0868 &66.62 &20.31\\
        $\checkmark$ &$\checkmark$ &$\checkmark$ & &0.893 &0.945 	&0.910 	&0.0292 & 67.59 & 21.66\\
        $\checkmark$ &$\checkmark$ &$\checkmark$ &$\checkmark$ &\textbf{0.912}  &\textbf{0.956}  &\textbf{0.922}  &\textbf{0.0203} &\textbf{68.41} &\textbf{24.98}\\
		\hline
	\end{tabular}
	\label{Ablation_study_of_CFLN}
	\vspace{-0.1cm}
\end{table}

\textbf{Complementarity of Loss Function.} Table \ref{Ablation_study_of_loss_function} presents a discussion on various loss functions, including $\mathcal{L}_{\text{consist}}$, $\mathcal{L}_{\text{BCE}}^{w}$, and $\mathcal{L}_{\text{IoU}}^{w}$. When using only $\mathcal{L}_{\text{consist}}$, the model produced the lowest scores, with $F_{\beta}^{w}$ at 0.514 and $M_{AE}$ at 0.2489. While semantic knowledge effectively supports saliency localization, it remains insufficient for precise segmentation of object boundaries. Adding $\mathcal{L}_{\text{BCE}}^{w}$ significantly enhanced performance, raising $F_{\beta}^{w}$ to 0.856 and lowering $M_{AE}$ to 0.1369. The model achieved optimal scores when all three loss functions were combined, suggesting that these loss functions complement one another to deliver the most efficient model performance across all metrics.
\begin{table}[!htp]\small
	\setlength{\abovecaptionskip}{0.2cm}
	\renewcommand\arraystretch{1.0}
	\tabcolsep 0.066 in
	\centering
    \caption{Ablation study of loss function terms.}
	\begin{tabular}{c|cc|cccc}
		\hline
		$\mathcal{L}_{consist}$ & $\mathcal{L}_{BCE}^{w}$ & $\mathcal{L}_{IoU}^{w}$ & $F_{\beta}^{w} \uparrow$ & $E_{\phi}^{m} \uparrow$ & $S_{\alpha} \uparrow$ & $M_{AE} \downarrow$ \\
		\hline
		$\checkmark$ &  &  & 0.514 & 0.775 & 0.656 & 0.2489 \\
        & $\checkmark$ & $\checkmark$ &0.897  &0.942 	&0.903 	&0.0355 \\
        $\checkmark$ &  & $\checkmark$ & 0.885 & 0.941 & 0.899 & 0.0378 \\
        $\checkmark$ & $\checkmark$ & & 0.856 & 0.925 & 0.873 & 0.0706 \\
        $\checkmark$ & $\checkmark$ & $\checkmark$ &\textbf{0.912}  &\textbf{0.956}  &\textbf{0.922}  &\textbf{0.0203} \\
		\hline
	\end{tabular}
	\label{Ablation_study_of_loss_function}
	\vspace{-0.2cm}
\end{table}

\textbf{Effectiveness of Consensus Deterministic Sampling.}
Table \ref{Ablation_study_of_CDS_sampling} presents a discussion on the impact of CDS scheme for saliency segmentation. Without CDS scheme (``-w/o CDS''), the model achieves lower scores on all four evaluation metrics. In contrast, with CDS scheme (``-w/ CDS''), the scores improved across all metrics, indicating a significant enhancement in segmentation accuracy and a reduction in errors. We further perform a visual comparison between the two cases. As shown in Figure \ref{4-4CDS_sampling}, it can be seen that the CDS scheme significantly improves the fine-grained segmentation performance of marine instances.

\begin{table}[!htp]\small
	\setlength{\abovecaptionskip}{0.1cm}
	\setlength{\belowcaptionskip}{-0.2cm}
	\renewcommand\arraystretch{1.0}
	\tabcolsep 0.086 in
	\centering
    \caption{Ablation study of Consensus Deterministic Sampling.}
	\begin{tabular}{cc|cccc}
		\hline
		-w/o CDS &-w/ CDS &$F_{\beta}^{w}\uparrow$ &$E_{\phi}^{m}\uparrow$ & $S_{\alpha}\uparrow$ & $M_{AE}$ $\downarrow$\\
		\hline
		$\checkmark$ & &0.903 &0.938 &0.916 &0.0267 \\
		&$\checkmark$ &\textbf{0.912}  &\textbf{0.956}  &\textbf{0.922}  &\textbf{0.0203} \\
		\hline
	\end{tabular}
	\label{Ablation_study_of_CDS_sampling}
	\vspace{-0.2cm}
\end{table}

\begin{figure}[!h]
	\setlength{\abovecaptionskip}{0.1cm}
	\setlength{\belowcaptionskip}{-0.2cm}
	\centering
	\includegraphics[width=0.46\textwidth]{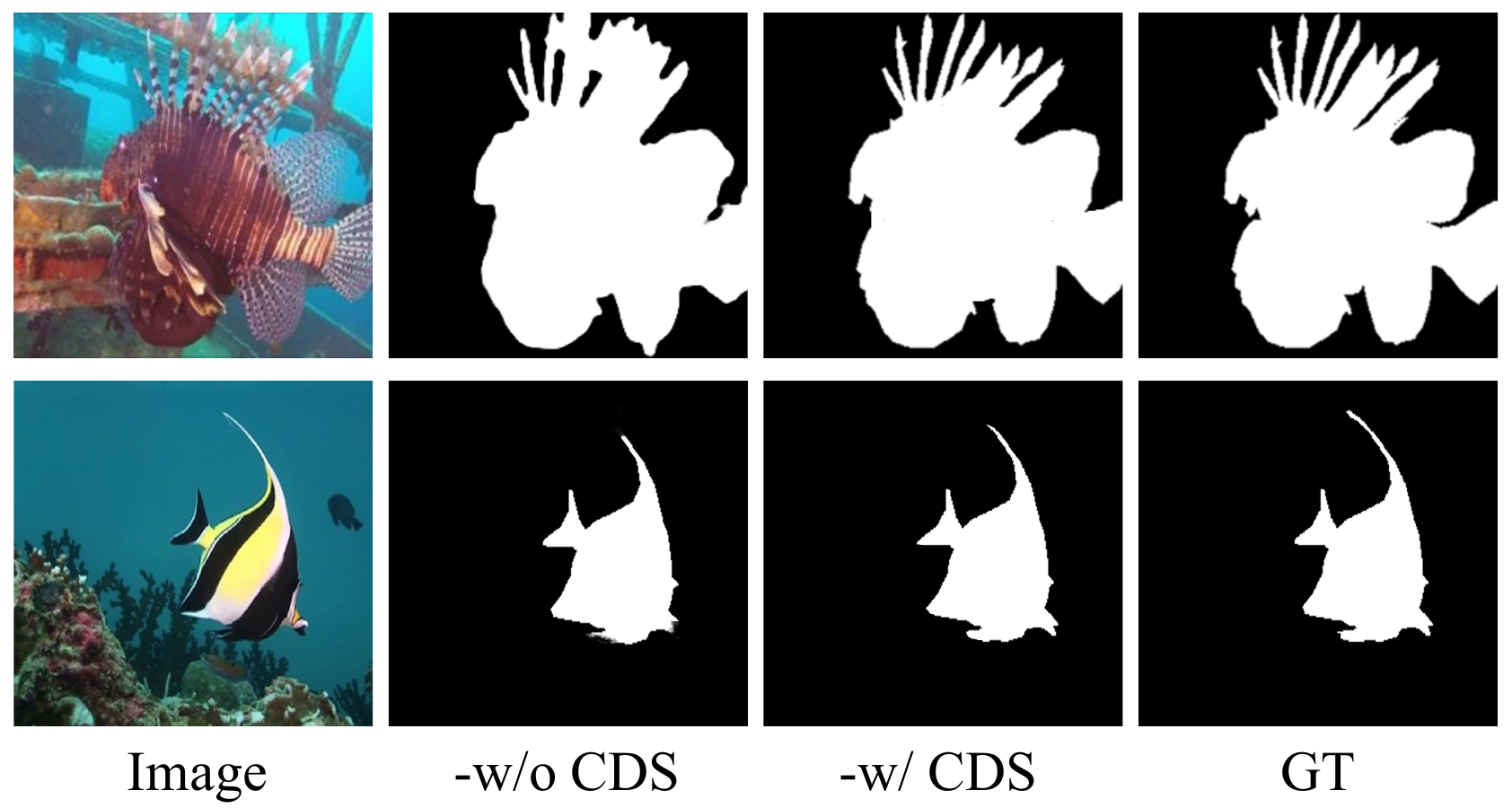}
	\caption{Visual ablation of Consensus Deterministic Sampling.}
	\label{4-4CDS_sampling}
	\vspace{-0.2cm}
\end{figure}

\section{Conclusion}\label{Conclusion}
In this paper, we present DiffMSS, an object-focused conditional diffusion model designed to leverage semantic knowledge distillation for segmenting marine objects. Our model introduces a region-word matching mechanism to enable word-level selection of salient terms. These high-level textual semantic features are then utilized to guide the CFLN module in generating diffusive conditions through semantic knowledge distillation. To further enhance segmentation accuracy, we propose the CDS scheme, which effectively suppresses missegmentations of objects with fine-grained structures. Extensive experiments validate that DiffMSS surpasses the state-of-the-art methods in both quantitative and qualitative results.

\small
\bibliographystyle{ieeenat_fullname}
\bibliography{main}

\end{document}